# A Framework of Full-Process Generation Design for Park Green Spaces Based on Remote Sensing Segmentation-GAN-Diffusion

Ran Chen[a,#], Xingjian Yi[a,#], Jing Zhao[a,*], Yueheng He[a], Bainian Chen[a],

Xueqi Yao[a], Fangjun Liu[a], Zeke Lian[a], Haoran Li[b]

**Abstract:** The development of generative design driven by artificial intelligence algorithms is speedy. There are two research gaps in the current research: 1) Most studies only focus on the relationship between design elements and pay little attention to the external information of the site; 2) GAN and other traditional generative algorithms generate results with low resolution and insufficient details. To address these two problems, we integrate GAN, Stable diffusion multimodal large-scale image pre-training model to construct a full-process park generative design method: 1) First, construct a high-precision remote sensing object extraction system for automated extraction of urban environmental information; 2) Secondly, use GAN to construct a park design generation system based on the external environment, which can quickly infer and generate design schemes from urban environmental information; 3) Finally, introduce Stable Diffusion to optimize the design plan, fill in details, and expand the resolution of the plan by 64 times. This method can achieve a fully unmanned design automation workflow. The research results show that: 1) The relationship between the inside and outside of the site will affect the algorithm generation results. 2) Compared with traditional GAN algorithms, Stable diffusion significantly improve the information richness of the generated results.

[a] School of Landscape Architecture, Beijing Forestry University, Beijing, 100083, China
[b] Director of China United Network Communications Co Ltd, Beijing, 100140, China
[#] Ran Chen and Xingjian Yi contributed equally to this manuscript
*Corresponding author

# 1.Introduction

Urban green space is an essential component of human living environment, and parks are among the most important elements of urban green space system. Park design is crucial for enhancing the livability, sustainability and service function of urban environment(Semeraro, Scarano, Buccolieri, Santino, & Aarrevaara, 2021). Therefore, park design requires a comprehensive consideration of various factors inside and outside the site. In traditional park design work, designers mainly rely on their professional knowledge and experience, resulting in low work efficiency and a limited number of design schemes for selection. In 2014, Lan Goodfellow(Goodfellow et al., 2020) proposed the generative adversarial network (GAN) technology, which has a strong generation capability and creativity, and can produce diverse and innovative schemes. This technology has significant implications for intelligent generation in design field. Based on GAN, park generation design optimization can greatly improve design efficiency, provide various high-quality design scheme previews quickly, support interactive exploration and rapid detail adjustment in early conceptual design stage, assist designers to complete high-quality design work, and bring practical benefits to urban green space planning and design projects.

Park design is a complex image generation task but a purposeful, comprehensive, and technical creative activity. It can be roughly divided into two categories: the first is to solve complex problems based on site constraints and generate design schemes accordingly; the second is to express the design schemes in standard drawings. There are some existing researches on generative design in these two tasks.

In the first category, Liu(Liu et al., 2022) et al. adjusted the layout label set continuously and guided the generation of Jiangnan private garden layout schemes that conform to the design logic. Chen(Chen et al., 2023)selected small and medium-sized park green spaces as the research object and constructed an automatic design system based on generative adversarial networks (GANs). However, most of the existing research on landscape architecture generative design ignores the influence of the external environment information of the site, which is particularly important in park design, on the design generation scheme. In the second category, Zhou(Zhou & Liu, 2021) et al. realized the rapid identification and rendering generation of land use classification without involving site layout generation with 325 finely annotated plan drawings. However, the generated results in generative design could be more precise and more accessible to present as the final scheme.

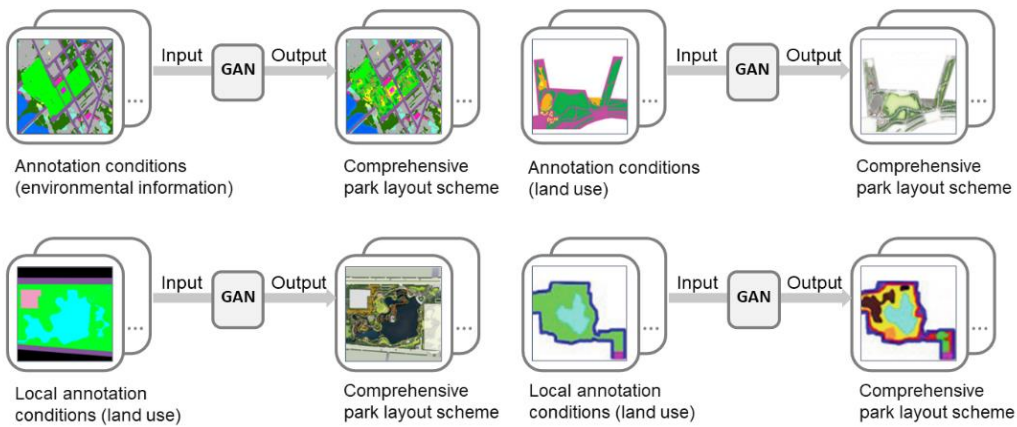

**Fig.1.** Design application demonstration based on GAN.

The landscape architecture plan should require more design details. The external environment of the site is one of the critical factors to consider in landscape architecture design. With the development of artificial intelligence algorithms, deep learning methods based on convolutional neural networks have been widely applied in computer vision, and their object recognition and classification techniques can be applied to the acquisition of environmental information. For example, high-resolution remote sensing images' recognition and semantic segmentation can extract various land features in remote sensing images. However, they have yet to be applied to extracting park external environment information (see Section 2.3).

To solve the above limitations and research gaps, this paper constructs a park generative design system based on GANs of "external environment-layout-scheme." The system intervenes in the two stages of design generation, controls the generation of the plane layout by setting practical constraints, renders the color texture map from the plane layout, and couples stable diffusion technology to optimize the details of the color texture map. The main contributions of this research are as follows.(1) Create an annotated small and medium-scale park green space design scheme dataset with park design element classification of "remote sensing-layout-scheme"; (2) Based on GANs, extract complex external environment elements of the site and complete the park green space layout generation based on multiple external environment conditions, significantly improving the "scientificity" of generation and the "reliability" of results; (3) Use stable diffusion to enhance the detail richness of rendered drawings and confirm its applicability to optimize the park generative design results; (4) By comparing different learning modes and different generative design constraints of neural networks, we verify that our generative design system performs better in generating high-quality and reasonable park layout schemes.

The remainder of this study is structured as follows: Section 2 reviews the related work on generative design and remote sensing extraction. Section 3 describes the proposed generative design system in detail. Sections 4 and 5 report and analyze the experimental results, respectively. Section 6 concludes the study based on the analysis and discussion.

## 2.Relevant work

This section is divided into three parts: the first part reviews the related literature on the application of GAN in planning and design; the second part discusses the current state of research on layout generation based on constraint conditions; and the third part briefly summarizes the methods of remote sensing feature extraction and explores the feasibility of extracting external environment information of the site as a control condition for generative design.

### 2.1 GAN-based generative design research

Generative adversarial network (GAN) was proposed by Ian Goodfellow in 2014(Goodfellow et al., 2020),and consists of two components: a generator and a discriminator. The generator tries to generate samples that will pass the discriminator's review, and the discriminator tries to distinguish synthetic samples from natural samples, competing with each other to form images similar to experimental data(Wu, Stouffs, & Biljecki, 2022). GAN's image creation generation ability far exceeds the previous deep learning models, so it has been applied more and more in the design scheme generation of human habitation environment(Zhao, Chen, & Bao, 2022). Based on GAN, generative design was first applied to the interior space layout generation of architecture. Huang(Weixin Huang & Zheng, 2018) and Zheng(W Huang, Williams, Luo, Wu, & Lin, 2018) used pix2pix to identify and generate architectural drawings, explored the layout of interior space plan, and established the main research paradigm in this field. Subsequently, this application method gradually entered the field of architectural design and urban and rural planning, mainly for automatically generating architecture and urban layou(Lin, 2020; Quan, 2022; Wu & Biljecki, 2022; Wu et al., 2022); at the same time, it was also applied to the rapid rendering of color texture map in urban and rural planning, such as Ye(Ye, Du, & Ye, 2022) et al. used GAN-based method to quickly generate high-quality rendered master plan images from vector CAD design line drawings. In outdoor space design, park green space has more complex space, more flexible layout, and more complex multidimensional design information in design drawings, which is difficult to achieve the same effect of rapid scheme generation by continuing the above methods. At present, the research paradigm of park green space generative design is generally limited to "site layout planning" (Chen et al., 2023), such as Liu(Liu et al., 2022) et al. adjusted the layout label set continuously and guided the generation of Jiangnan private garden layout schemes that conform to the design logic; Zhou et al. used GAN to identify and generate landscape plan; Chen(Chen et al., 2023) et al. based on pix2pix and CycleGAN realized the generation of park green space spatial layout and the rendering of color plan.

In summary, there is little research on GAN in park generative design. Its generative design method continues the generative design method of architecture and urban and rural planning. The optimization of its control mode is also through multiple experiments, through trial and error of various site internal element layout labels to determine the control mode, which has not yet considered other design influencing factors on its generation constraints.

## 2.2 Generative design based on constraint control

Constraint-based scheme layout generation was first applied to small-scale indoor space design. Initially, the constraints were limited to the inside of the red line. Zheng(W Huang et al., 2018) et al. applied CGAN technology to connect the design boundary as the architectural interior design constraint and the plan, and introduced the annotation of room type and architectural axis, which enhanced the reliability of the generated results. Huang(Weixin Huang & Zheng, 2018) et al. applied Pix2PixHD technology to further refine the constraints, and annotated nine types of room types and details such as doors and windows, making the indoor layout more clear and reasonable. Wang(S. Wang et al., 2021) et al. simulated the generation of indoor pedestrian trajectories by bi-RRT algorithm and used them as constraints, along with the architectural design scope, to design the indoor design generation under pedestrian trajectory constraints. Wang et al. developed a new deep learning framework ActFloor-GAN, which uses human activity maps to guide the generation of architectural floor plans from input boundaries. Their research considers the design conditions in natural design from the designers' perspective and provides possibilities for the algorithm to explore multiple design schemes based on constraint conditions.

With the development of related research in the outdoor design field, researchers tried to extract site external environment information as constraint conditions for scheme generation. Lin Wenqiang(Lin, 2020) et al. based on Pix2Pix technology, completed the primary school campus layout generation with design red line and urban road as constraint conditions. The urban road constraint contains more semantic information of design, which enhances the rationality of outdoor design scheme generation. Liu(Liu, Luo, Deng, & Zhou, 2021) et al. completed the rapid generation of university campus layout scheme in two steps: first, based on external environment to generate functional zoning; then, based on functional zoning to generate architectural layout. Stanislava Fedorova(Fedorova, 2021) used GAN technology to achieve similar applications, using urban surrounding environment as learning training data, and based on the surrounding texture to complete the blank area.

Constraint-based scheme layout generation has been gradually applied to medium-scale outdoor layout scheme generation, which has application potential in park production design research. However, the research based on the external environment as a constraint condition is limited, and it is challenging to meet the requirement of park design to extract complex environmental elements as constraint conditions.

*2.3 Remote sensing object extraction*

Deep learning methods based on convolutional neural networks have been widely applied in computer vision, and their target classification, recognition, and tracking technology have reached a high level. They have been applied to high-resolution remote sensing image recognition and semantic segmentation. The applications in forestry, construction, water conservancy, and other fields are mainly based on extracting single-class objects, such as plants(Xie et al., 2022), building(Zhang, Zhang, Liu, Qiao, & Wang, 2023), roads(Linghu, Xiping, Shu, Lin, & Mingyu, 2023), water(H. He et al., 2020), etc. In recent years, many studies have improved these applications and achieved rapid extraction of multiple land types from remote sensing images. For example, Lu et al. used Deeplab v3+ segmentation model for high-precision classification of cultivated land, vegetation, buildings, roads, and water systems(Lu, Mao, He, & Song, 2022); Tzepkenlis used an improved U-Net method to segment satellite images into multiple land cover types, with performance significantly higher than traditional deep learning methods(Tzepkenlis, Marthoglou, & Grammalidis, 2023). However, problems such as loss of detail information, incomplete segmentation and large parameter size are still prominent. To this end, some researchers tried to combine GAN technology to extract objects from remote sensing images. In He Ping(He, Zhang, & Luo, 2020) and Wang Yulong(Y. Wang, Pu, Zhao, & Li, 2019) et al.'s research, it was found that GAN method had better segmentation effect on small-scale remote sensing images in urban areas. At the same time, Hamideh et al. addressed the problem of insufficient labeled data sets by using a semi-supervised framework based on GAN to train on a small amount of data sets, which also improved the performance of urban scene segmentation(Kerdegari, Razaak, Argyriou, & Remagnino, 2019). Therefore, GAN has some advantages in semantic segmentation processing of small-sample remote sensing data.

The remote sensing image pattern of the urban park's external environment is complex and challenging, and the data volume is small. If traditional semantic segmentation models are used, it is easy to overfit. This study uses remote sensing image object extraction based on GAN to provide external environment conditions for generative design and realize controlled generation based on the external environment.

## 3.Methodology

This study aims to propose a park layout scheme generation system based on an external environment, which can quickly generate park layout schemes and color plans and provide design inspiration for designers. The overall experimental framework includes four steps(see Fig.2): (1) data set creation; (2) object classification extraction of urban remote sensing images; (3) park design scheme generation based on the external environment; (4) detail optimization of a design scheme. The first step is to collect and create data sets with semantic annotations and graphic structures labeled urban remote sensing object information data sets and park design scheme data sets. The second step is to use the remote sensing object information data to train CycleGAN, construct the object information extraction method, and use this method to obtain the park's external environment information. The third step is to use the park's external environment as the constraint condition and send it into pix2pix and CycleGAN, two neural networks for image generation, to train and generate park layout schemes and design schemes. The last step is to introduce Stable Diffusion technology to optimize and complete the details of the design scheme generated by the algorithm and discuss the application mode and difficulties of the AIGC algorithm model in the landscape architecture field. The following sections will provide more details about each step.

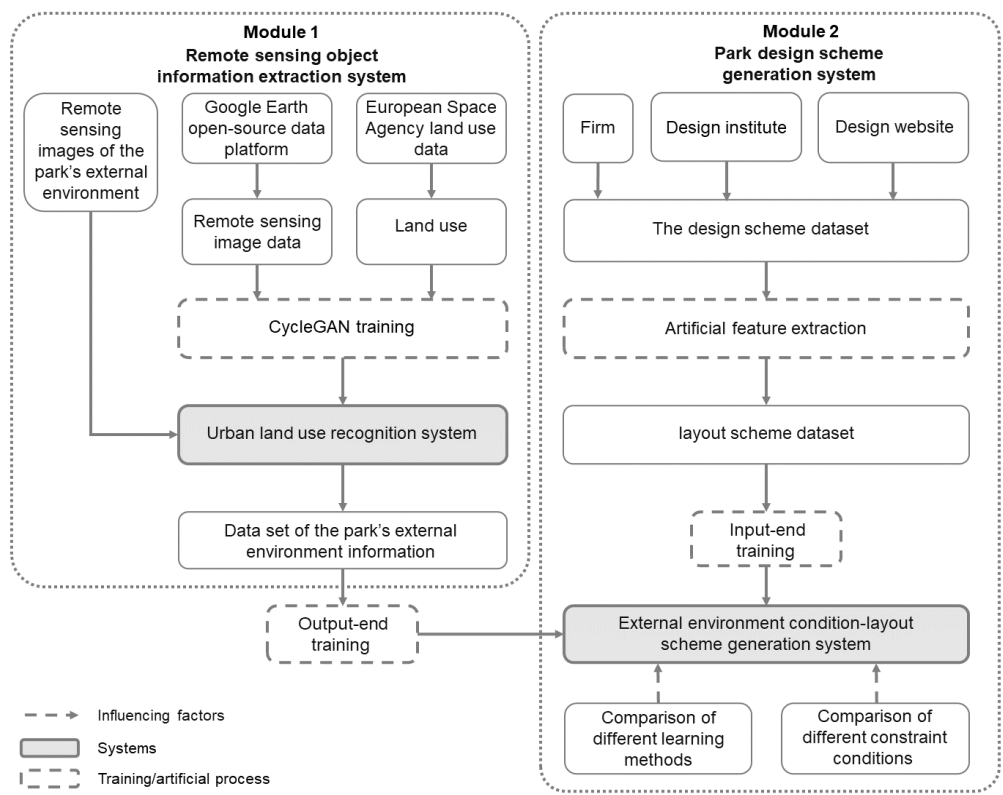

**Fig. 2.** Methodology for automated park design schemes generation based on external environment condition.

*3.1 Data set creation*

This step involves creating two data sets: remote sensing object information data and park layout scheme data. The CycleGAN algorithm requires paired data for unsupervised learning, while the Pix2Pix algorithm requires one-to-one annotated data sets. Therefore, this study will process the data sets into images of the same size and format.

### 3.1.1 Remote sensing data

This part of the data consists of the remote sensing images and land use information of five blocks of the same size in the Beijing area, which are obtained from Google Earth and ESA global land use data.

To improve the accuracy of the object extraction system, the selection of training data follows two principles: (1) medium and small-scale green space: large-scale green space will cause too much compression of image information in the image processing process of deep learning, making it challenging to identify the scheme details; (2) strong regularity: deep learning is an algorithm model based on probability statistics, and its input data needs certain regularity. For example, in urban areas, the combination forms of squares, roads, and green spaces are diverse, and some urban areas have few buildings and hard surfaces with weak regularity. Therefore, to improve the training efficiency and ensure that the land use and remote sensing images in the training data are consistent in size, this step uses Python language and Opencv tool kit to cut them into small pieces of 512*512 pixels, with a resolution of 5 meters/pixel. Finally, 1000 remote sensing images and land use images were obtained as training data for remote sensing object extraction.

### 3.1.2 Park data

This study focuses on medium and small-sized parks, including community parks (1-10 hectares), a few small-scale comprehensive parks (more than 10 hectares), and other green space types that meet the requirements. The data includes park external environment remote sensing images and park layout schemes.

TTo avoid the failure of a deep learning algorithm in complex and diverse outdoor space data, we need to select high-quality data samples, considering both the algorithm's understanding mode and the characteristics of park design. This experiment selects experimental samples based on four aspects: (1) strong regularity, (2) less special design conditions, (3) clear functional zones, (4) slight land elevation difference. Finally, we selected 137 medium and small-sized parks as training samples and 6 test samples. The test samples include as many functional zones and common park layout patterns as possible, providing references for generating complex spatial types. The accurate annotation of park layout elements is the key to obtaining good algorithm training results. Chen(Chen et al., 2023) et al. determined eight significant categories of park layout key elements by semantic segmentation and comparison of different styles of park planning: green land, water, road, paving, red line, urban road, and plant. This study refers to this classification method and divides the park plan into six categories: green land, water, road, paving, building, and plant(see Table 1). Among them, water bodies and buildings are extracted separately as site information.

**Table 1**
Land use code table.

| Land use | Code(R,G,B) |
|---|---|
| **Green land** | (0,255,0) |
| **Water** | (0,255,255) |
| **Roads** | (241,145,73) |
| **Paving** | (255,255,0) |
| **Structures** | (255,0,255) |
| **Plant** | (0,152,67) |

We selected 137 small and medium-sized parks that met the requirements and obtained their remote sensing images, layout schemes, and base information(see Fig.3). We used the remote sensing images as test data for information extraction and the layout schemes and base information as training data for rapid generation of park layout schemes.

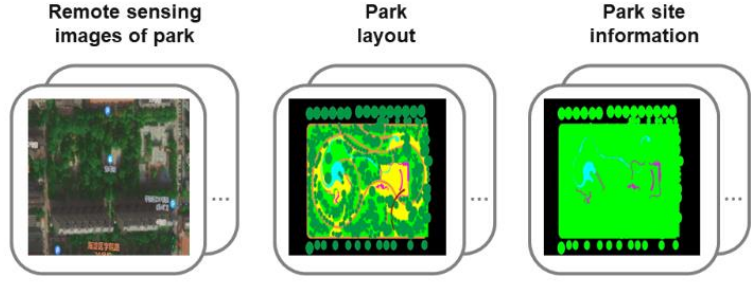

**Fig. 3.** Display of three types of park data.

*3.2 Object classification extraction of urban remote sensing images*

This study proposes an object information extraction method based on the reliable records of previous research on deep learning technology. It uses data sets that conform to the characteristics of the park's external environment remote sensing to train to improve the accuracy and reliability of the park's external environment information. Obtaining ideal results from small-sample data training algorithms is an urgent problem to be solved. From remote sensing images to remote sensing object classification color block images, similar to style transfer tasks, the CycleGAN algorithm is one of the most widely used algorithms in this kind of task. Therefore, this study uses the CycleGAN algorithm to extract remote-sensing object information.

CycleGAN is a generative adversarial network (GAN), proposed by Zhu(Zhu, Park, Isola, & Efros, 2017) et al. in 2017, applied to unsupervised image generation. Its outstanding feature is that the input data does not need paired data; it only needs to provide two different types of data that can be used for image generation. CycleGAN has two built-in generators, and the overall framework is shown in [Figure 1]. Taking remote sensing object extraction as an example, assuming that there are two domains X and Y, X can be understood as object information color block diagram, Y as remote sensing image. CycleGAN has two generators: G and F. Generator G is used to generate Y-domain images based on X-domain images (color block diagram → remote sensing image), while generator F is used to generate X-domain images based on Y-domain images (remote sensing image → color block diagram). These two generators have opposite and independent functions, which loss functions can restrict. At the same time, CycleGAN has two discriminators, DX and DY, which are used to judge whether the input X-domain or Y-domain images are real. The overall network structure can be seen as a combination of two GANs, forming a cyclic process that can use loss functions to constrain the relationship between generated images and input images in the opposite generation process.

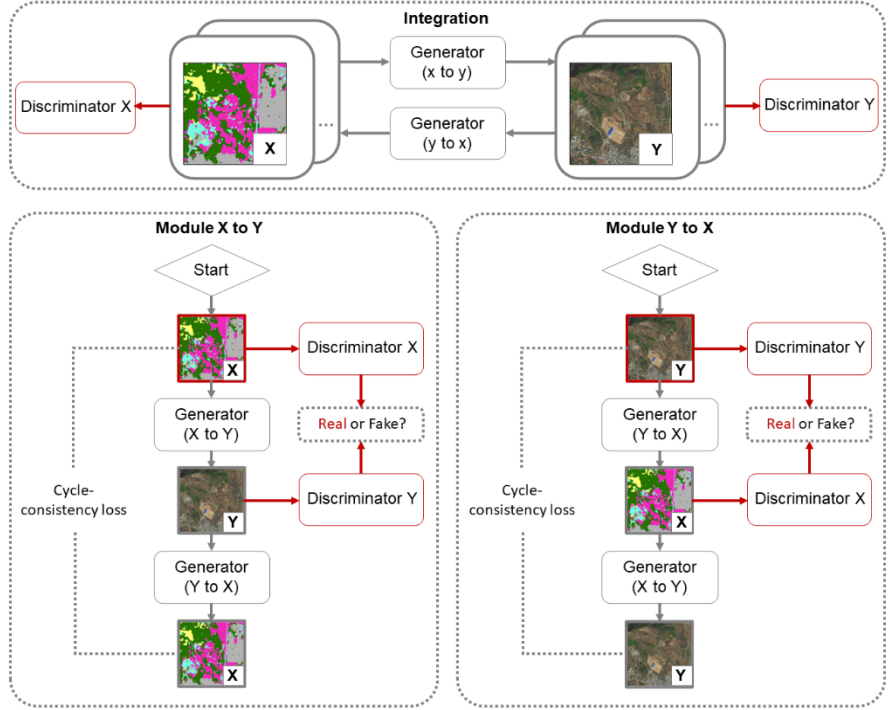

**Fig. 4.** Diagram of the CycleGAN model.

*3.3 Park design scheme generation based on external environment*

This generation process consists of two parts: (1) the generation process from the external environment information color block diagram to the park layout scheme color block diagram, and (2) the generation process from the park layout scheme color block diagram to the park design scheme plan. The original GAN model focuses on clear and regular spatial design and pays less attention to complex layout generation problems. To address this problem, Chen et al. proposed LandscapeGAN based on pix2pix and CycleGAN and constructed an automated design generation system for park green space. However, the layout generation based on external environment constraints is similar to the image completion task, where the algorithm uses the image itself or image library information to complete the missing area of the image to be repaired, rather than the style transfer task that the GAN model often performs. Therefore, this study makes improvements based on LandscapeGAN.

In this algorithm model, the technical principle of Cycle GAN is the same as that in Section 3.2. The input end of the training data is park external environment information data, which is predicted by the algorithm in Section 3.2. The output end of the training data is manually drawn layout scheme data. The pix2pix algorithm is a kind of algorithm based on GAN to realize image mapping, which was proposed by Isola(Isola, Zhu, Zhou, & Efros, 2017) et al. in 2017. It is slightly different from the original GAN model in terms of image transformation strategy. It is slightly different from the original GAN model regarding image transformation strategy. Its core idea is to use the input image as conditional information to guide the learning of the generator and discriminator. The goal of the generator is to generate realistic output images based on input images, while the goal of the discriminator is to distinguish authentic output images and generated output images and give corresponding scores.

The basic structure of the Pix2Pix algorithm includes a generator and a discriminator (see Fig.5). The generator uses the U-Net architecture, which consists of an encoder and a decoder. The encoder compresses the input image into a low-dimensional feature vector, and the decoder restores the feature vector into a high-dimensional output image. The discriminator uses the PatchGAN architecture, which judges local areas of images and gives true-false scores for each area. However, this algorithm requires paired data for training and may produce some blurred or distorted output images.

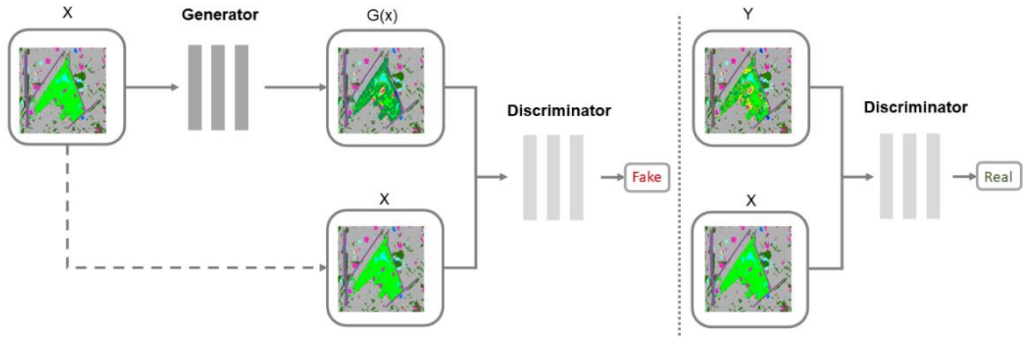

**Fig. 5.** Diagram of the pix2pix model.

*3.4 Detail optimization of design scheme*

The above process has completed the generation process from external environment conditions to the design scheme. However, there are many problems, such as unclear edges and noise in the design scheme generated by the algorithm, which are challenging to be presented as the final result of the park design scheme. Therefore, this study introduces the existing AIGC algorithm model based on Stable Diffusion to complete the details of the plan generated by the algorithm.

Stable Diffusion is an image generation model based on the diffusion process, which was released in 2022. The diffusion Model (DM) is a generative model based on the Markov chain(Ho, Jain, & Abbeel, 2020). It gradually adds Gaussian noise to input data, turns it into pure noise, and then uses a neural network to reverse denoise and reconstruct data from noise. Its advantage is that it can generate high-quality images, but its disadvantage is its slow generation speed and requires multiple iterations.

The stable Diffusion Model (SDM) uses a stable latent variable z based on the latent diffusion model, which no longer adds noise with time step but uses a deterministic function to update z. It encodes data into z by an encoder and updates z at each time step using an update network, then decodes z into data by a decoder. Therefore, the Stable Diffusion model can further improve generation speed and quality. The overall framework of this model is shown in [Fig. 6].

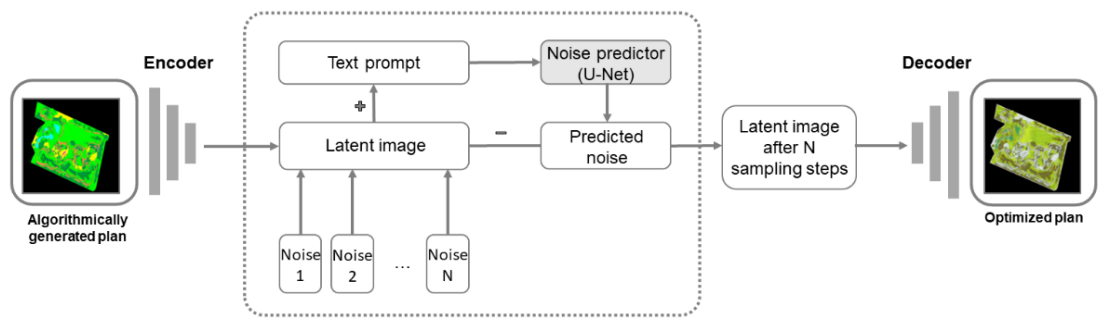

**Fig. 6.** Diagram of the SDM.

# 4 Experiment

This part shows the implementation and experimental settings of the process generation method from "remote sensing image-external environment information-park layout scheme-park design plan-plan optimization".

## 4.1 Implementation of GAN-based "external environment-layout-scheme" generation method

It consists of three steps: "remote sensing-external environment," "external environment-plan," and "plan optimization."

Firstly, we extract park external environment conditions based on remote sensing images, train with the Cycle GAN algorithm, and test to obtain park external environment information. The data training and testing process of the algorithm is as follows:

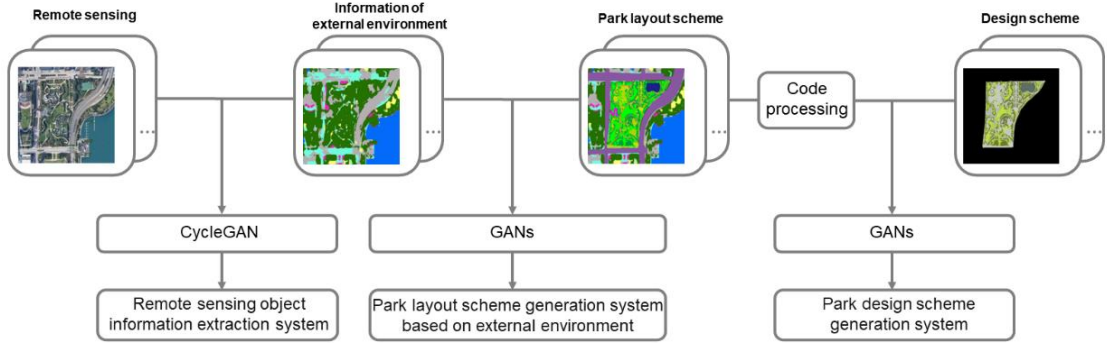

**Fig. 7.** The overall workflow of training and predicting.

Next, we use "external environment information-park layout" and "park layout-design scheme" as training data, train through Pix2Pix and CycleGAN networks, respectively, and construct a park layout generation system and park design scheme system based on the external environment.

Finally, we input the CycleGAN generated design plan into Stable Diffusion for optimization. According to pre-experiment results, the more significant proportion of the plan in the picture occupies, the better optimization effect will be obtained; therefore, we enlarge the design scheme in the data picture; this step uses PS to complete; at the same time, we input concise description words for assistance: "urban park, top view." Stable Diffusion does not need to train algorithms and can obtain generation results by directly inputting data.

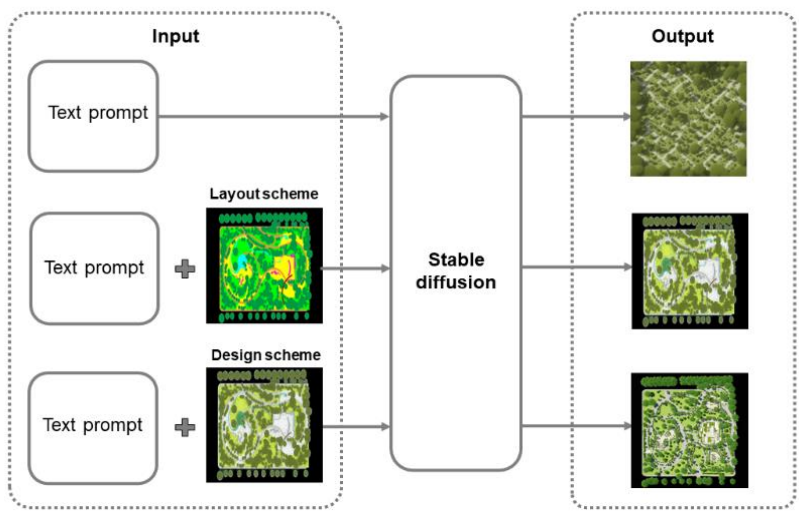

**Fig. 8.** The optimization of generated design plan.

## 4.2 Comparative experiments

This study conducted four groups of comparative experiments to examine the differences between different methods in various application scenarios in detail and described the application characteristics, advantages, and limitations of each method. The experiments are numbered from 1 to 9 according to the "external environment-layout-scheme generation process." The detailed information of the experiments is shown in [Table 2].

FFor the part of park layout scheme generation based on external environment conditions, experiments 1 and 2 compare the influence of different methods on generation effect by selecting different methods to study the learning mode; experiments 2 and 3 use different training data to study the influence of data set information on generation results, to determine the effect and necessity of external environment information as generation constraint condition; for the part of park design scheme generation based on park layout, experiments 4, 5 and 6 use different quality training data to compare the influence of different data quality on generation results; for the part of park design scheme optimization, experiments 6, 7 and 8 input text, layout scheme and text, design scheme and text respectively, to explore the application method of Stable Diffusion in park generation design, and analyze the method's understanding mode of design scheme.

**Table 2**
Experimental details of comparative experiments.

| Index | Experiment Name | Training Data at the Input End | Training Data at the Output End | Source of Data | Number of Dataset | Algorithm/ Software |
|---|---|---|---|---|---|---|
| E1 | Supervised learning training on external environment | External environmental information | Layout scheme | Source of external environmentdata: output data from module 1/ Source of layout scheme information data: manually drawn | 137 | Pix2pix |
| E2 | Unsupervised learning training on external environment | | | | | CycleGAN |
| E3 | Training on park interior environmental information | Interior environmental base information | | | | CycleGAN |
| E4 | Generating design proposals based on textual descriptions | Textual Description | Refined design scheme | Source of layout information data: manually drawn/ Source of design scheme information data: output data from experiment 2 | \ | Stable Diffusion |
| E5 | Generating design proposals based on layout scheme | Layout Scheme | | | | Stable Diffusion |
| E6 | Generating design proposals based on design scheme | Design scheme generated by Cycle GAN | | | | Stable Diffusion |

# 5.Result and Discussion

## *5.1 Remote sensing object information extraction stage*

To intuitively illustrate the experimental results of remote sensing land cover extraction using CycleGAN, Figure 9 shows some of the training data. The remote sensing images of the external environment of the park are the test set, which are input into the trained algorithm network for testing. Figure 11 shows the test results of six remote sensing images of the external environment of the park with representative land cover features to observe the network's ability to recognize remote sensing land cover.

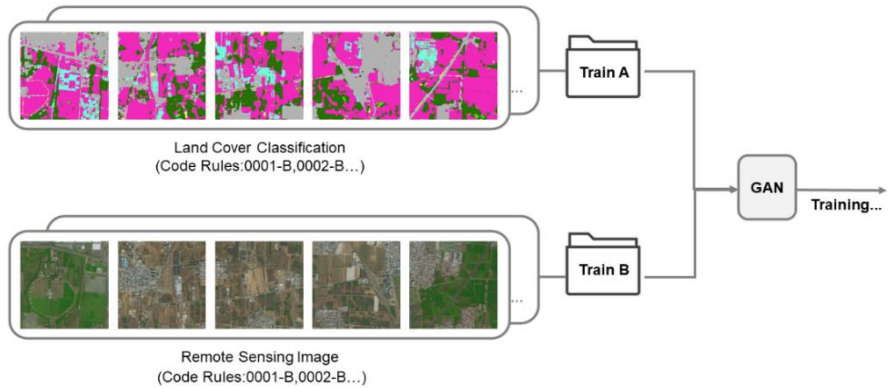

**Fig. 9.** Examples of data from label set (top) and feature set (bottom).

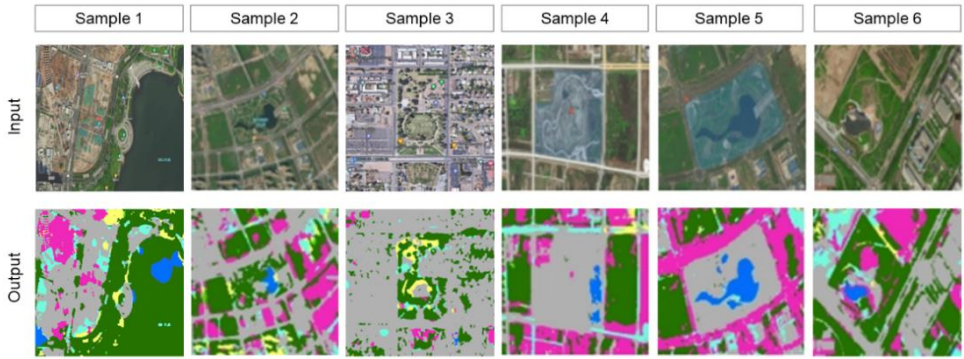

**Fig.10.** Partial results of remote sensing objects extraction.

The overall results show that the CycleGAN algorithm has a good classification effect on the remote sensing object information, and the generated feature classification method is mainly consistent with the input land use data, with less confusion, higher clarity, and few noise points.

a. Remote sensing feature classification is more accurate. In the remote sensing feature information color block diagram generated by the CycleGAN algorithm, the color block color corresponding to each feature is mostly the same as the input data. There is some confusion in the extraction of remote sensing image information, which may be due to some complex or blurred data in the training images. This suggests that under small sample training, the CycleGAN algorithm can accurately judge the characteristics of different features. However, the algorithm cannot summarize its characteristics accurately for features with complex textures and particular or diverse forms, resulting in deviations.

b. Features with similar colors that are easy to confuse. In most generated results, the CycleGAN algorithm failed to distinguish between urban roads, hard ground, and some concrete buildings. This may be because the colors of urban roads, hard ground, and buildings are similar and lack texture features. Moreover, in the city, the common forms of these three features are irregular, so the algorithm is complex to extract features. To solve this problem, more accurate land use data and remote sensing images with more uniform and apparent features of each land use need to be used for training to improve the training effect.

*5.2 Park design scheme generation stage based on external environment conditions*

### 5.2.1 Generation results of park layout scheme and design scheme

Based on the park data set, the design layout and scheme generation systems were trained in turn.The following figure shows the park layout scheme and design scheme trained and generated by CycleGAN, and the test samples are the external environment information extracted by CycleGAN and the layout scheme generated by CycleGAN.

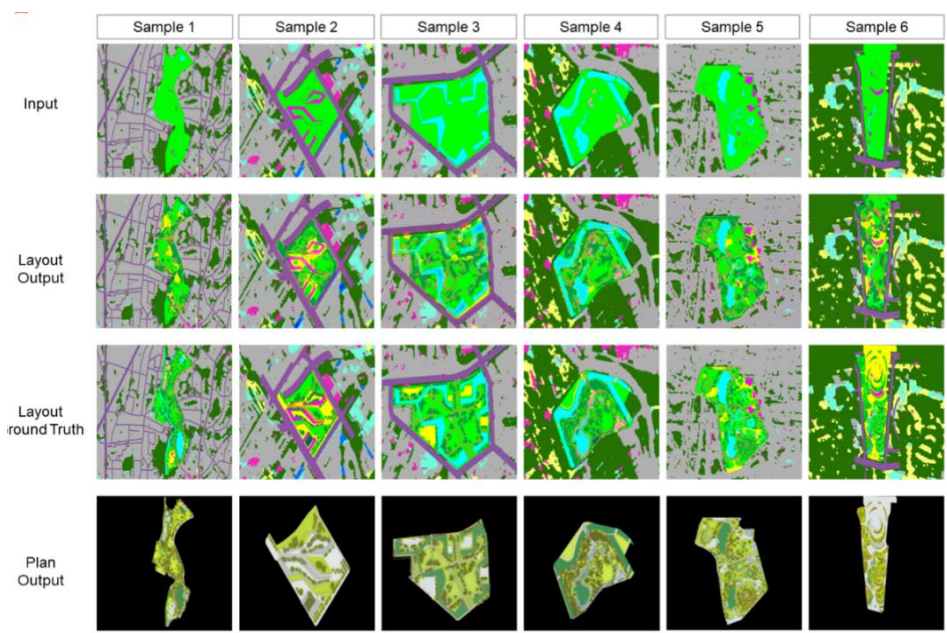

**Fig. 11.** Examples of park layouts and design plans generated by CycleGAN under the constraints of external environment.

Generally speaking, the CycleGAN algorithm can summarize the characteristics of various design elements to some extent, build the connection between structures and roads, paving, and other landscape elements, and reconstruct them according to the given environmental conditions. The detailed analysis of the generation effect is discussed in section 5.2.2.

### 5.2.2 Generation result evaluation

From the park data, we chose 50 cases that exhibit diverse and distinctive park layout features as our test samples to observe the applicability and accuracy of the trained network in different scenarios. The study also compared the layout scheme generation effects under different learning methods and constraints and the design scheme generation effects under different training data quality.

(1)The influence of the learning method on the layout scheme generation effect

Currently, pix2pix or CycleGAN are mainly used as the training algorithms for automatic generation research in the field of planning and design; Liu(Liu et al., 2022) , Zhou(Zhou & Liu, 2021), Ye(Ye, Du, & Ye, 2022), and others have used them. However, few studies compare and analyze the application scenarios of the two algorithms. This part compares the supervised learning training based on the Pix2Pix algorithm, and the unsupervised learning training based on the CycleGAN algorithm and tests the generation effects under different methods. The test set is input into the Pix2Pix algorithm, and CycleGAN algorithm trained in experiments 1 and 2, respectively, for testing, and the generation results are shown in Figure 12.

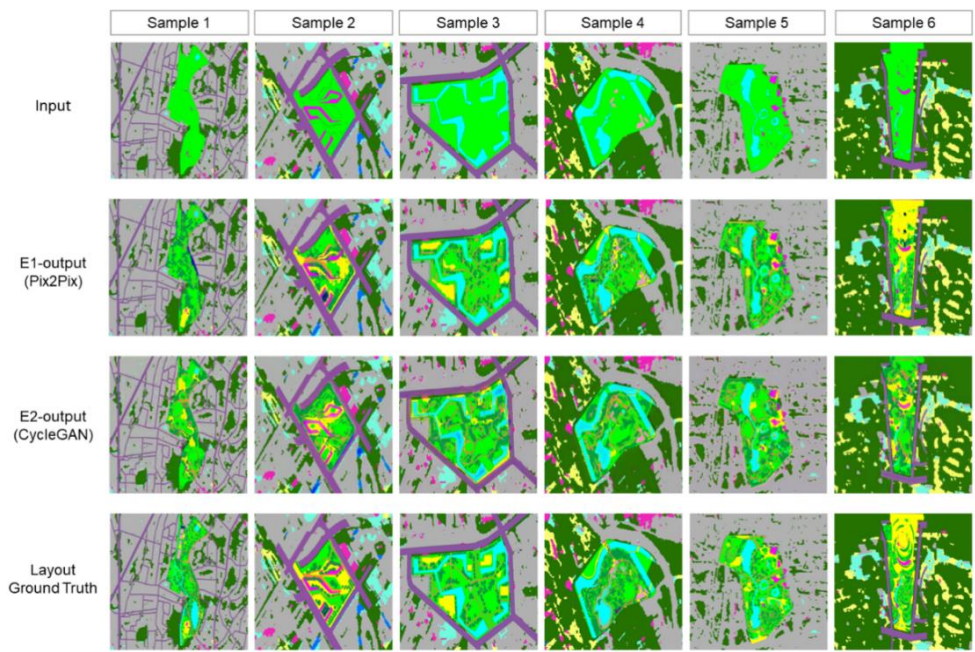

**Fig. 12.** Comparison of park layouts generated by different algorithms under the constraints of external environment.

In the test results of Pix2Pix, the algorithm can extract the features of different design elements and organize them with different colors. Its results are closer to the reference scheme than CycleGAN, but also show some defects: a. Some collapse cases occur, generating repetitive color block textures (such as Sample 1). b. Some layout elements have blurred boundaries, such as squares and roads, with more noise points in their contours (such as Sample 2 and Sample 3).

The test results of CycleGAN are relatively stable, and the layout and organization of different elements, such as roads, plants, and squares, are more reasonable and rich. CycleGAN can reasonably organize the spatial distribution according to the external environmental conditions, forming a spatial layout different from the original scheme, with a certain degree of innovation, showing the algorithm's design understanding of landscape elements. However, there are also some problems: a. When generating large areas of vegetation, there will be bluish areas or contours similar to water bodies (such as Sample 3 and Sample 4); b. Some roads are not coherent or clear (such as Sample 2).

The two algorithms have certain advantages and limitations in the park layout generation. Compared with CycleGAN, Pix2Pix performs better in the case of small sample training and highly restores the training data, but the generation results are difficult to unify in style; CycleGAN has a stable generation style and is more innovative and suitable for landscape design without standard answers. The defects shown by the two algorithms are mainly related to the low clarity and insufficient data, resulting in repetitive or noisy generation results. Based on the experimental analysis of this section, CycleGAN is better than pix2pix in terms of the stability and accuracy of the target generation, and has more potential in generative design.

(2)The influence of constraint conditions on the layout scheme generation effect

The test set is input into the CycleGAN algorithm trained in experiment 3 for testing and compared with the test results of experiment 2. The following figure shows some of the generated results. The study also evaluates the generation results of the two experiments from the aspects of the road system, spatial relationship, and design element understanding.

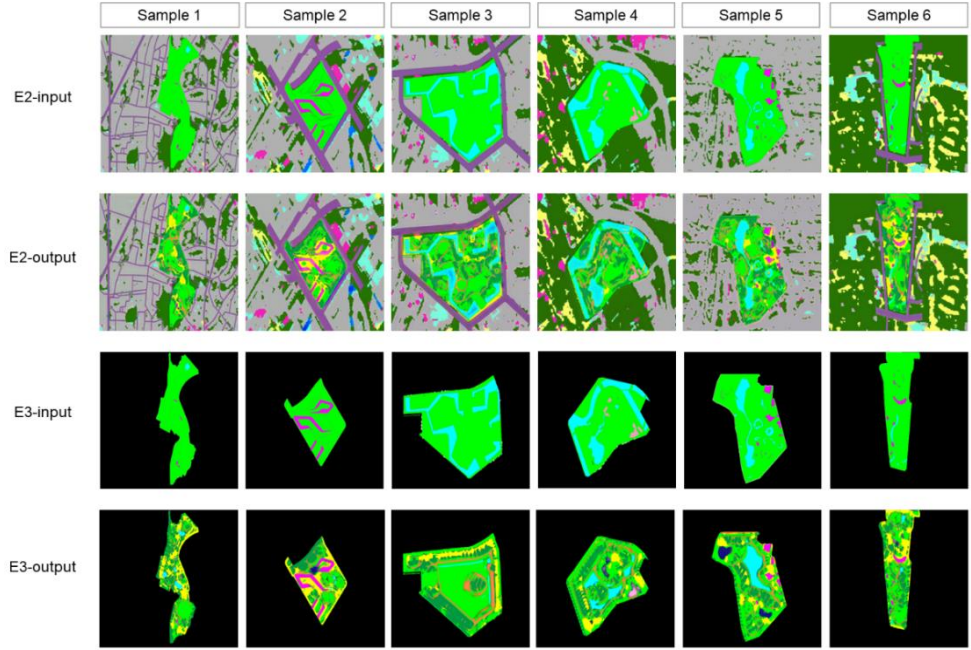

**Fig. 13.** Comparison of park layout generation results based on different constraints.

a. Road system aspect: In the generated results, the generated results of both experiments can form a park road design that runs through the whole park, and other design elements are closely combined, and basically, each paving node can be connected by the park road, providing a diverse and rich spatial experience.

When only the internal environmental information of the site is input as a constraint condition, there are some common problems in the generated results, such as the park roads in cases (a2) and (d2) repeat in loops, and in case (b2) some park roads keep a certain distance from the park boundary. However, the generated results with the internal and external environment of the park as the constraint conditions can form a more reasonable and smooth park road streamline, and the generation performance is more flexible and diverse and will be affected by the external environmental information. For example, in case (b1), the park road changes the density according to the number of other design elements, dividing the primary and secondary lawns, and at the same time, (b1) and (c1) have roads leading to the external city roads of the park, forming the entrance and exit space.

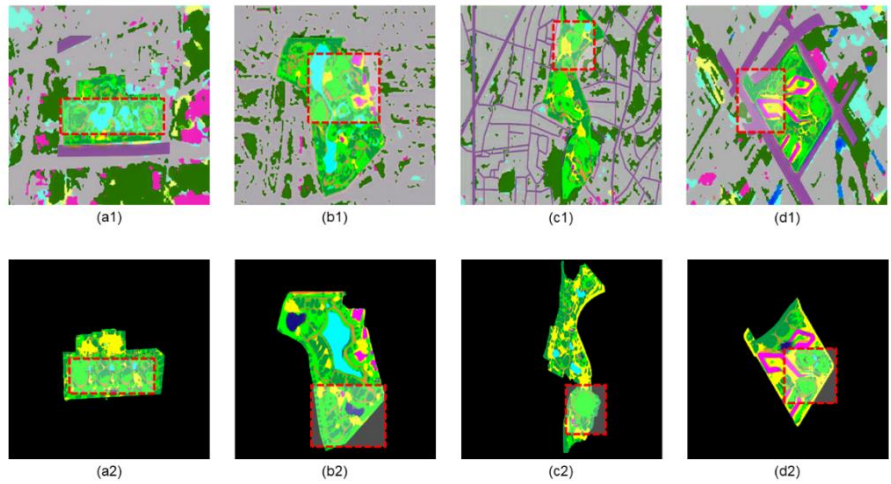

**Fig. 14.** Comparison of the road system generated by inputting environmental conditions inside and outside the park (top) and only internal conditions (bottom)

b. Spatial relationship aspect: The test results can form a diversified and harmonious space. When the environmental information inside and outside the site is used as a constraint condition, the algorithm can use various landscape elements more flexibly and variably and also consider the impact of the external environment on the park design, making the generated results closer to the actual park's spatial layout mode.

When only considering the internal environmental conditions of the site, some generated results will form multiple similar spaces locally, most of which are concentrated in the middle of the park, such as cases (e2) and (f2), which are slightly insufficient in creating the spatial experience. However, when considering the environmental information inside and outside the site at the same time, the generated park layout will connect the large-scale open space with the challenging square and the external city road of the park, such as (e1) forming a distribution space similar to the entrance square; in cases (f1) and (g1), most of the park boundary areas are divided by narrow forest belts, which reasonably separate the internal space of the park from the city; more semi-open or closed spaces are placed in the middle of the park, such as cases (f1) and (h1) where more semi-open or closed spaces are arranged in the central area of the park, creating a forest underpass experience.

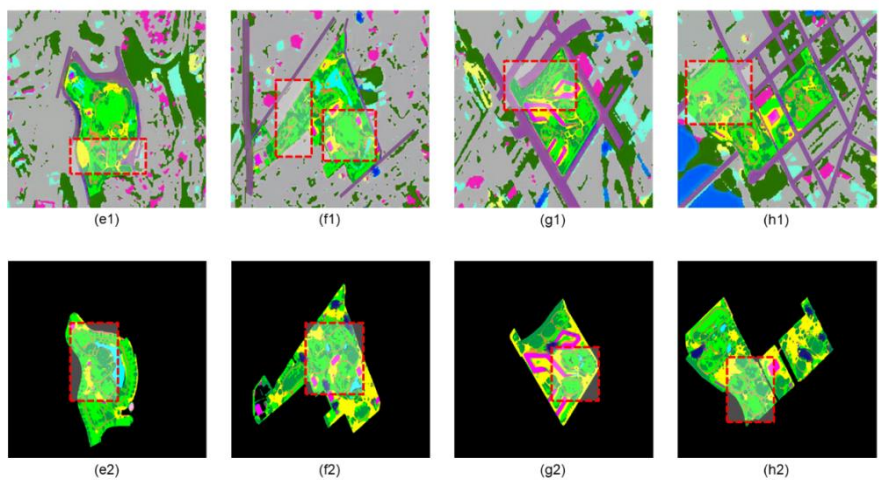

**Fig. 15.** Comparison of the spatial relationship generated by inputting environmental conditions inside and outside the park (top) and only internal conditions (bottom)

c. Design element understanding aspect: CycleGAN can roughly summarize the features of various design elements and build the connection between structures and roads, paving and other landscape elements.

In this round of experiments, the test results of inputting internal environmental information have a weak understanding of the features of structures and other elements such as paving, roads, etc., such as the case (i2) where most of the hard paving is concentrated in the middle of the park, or appears in a large area at the edge of the park, the east side road of case (j2) runs through the park in a straight line, and even appears chaotic. These all reflect that the algorithm-generated results lack consideration of the external environment of the park, resulting in a large gap between its layout mode and the actual park. The test results of adding external environmental information can better integrate the landscape nodes formed by structures and paving with the road system, such as case (i1) where a specific area of paving is generated around the two larger structures in the west and south as a connection, (k1) where a small square is generated between the north water body and the park boundary, and several spaces are connected by the park road.

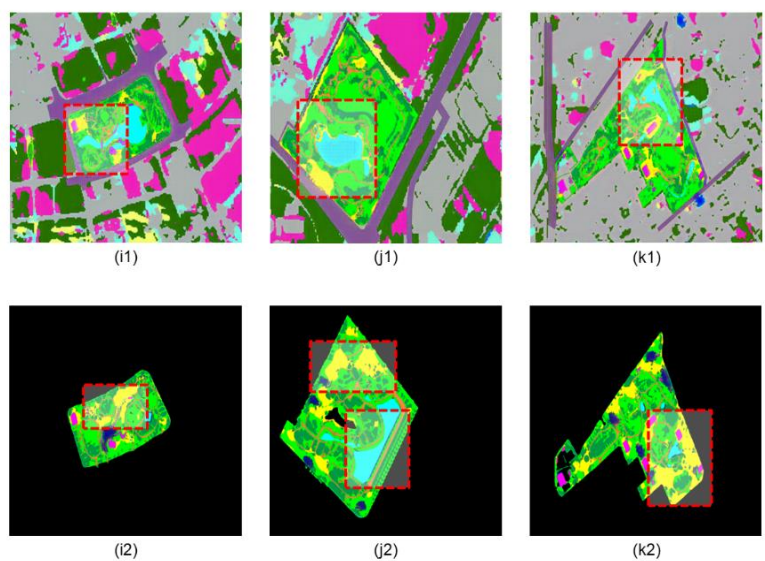

**Fig. 16.** Comparison of design elements generated by inputting environmental conditions inside and outside the park (top) and only internal conditions (bottom)

*5.3 Park design scheme optimization stage*

This part of the experiment aims to use the Stable Diffusion model to generate park design schemes, complete the details of the park design schemes generated by CycleGAN, and compare the generation effects of park design schemes with different data information input to Stable Diffusion.

a. Input text, difficult to control the generation results.

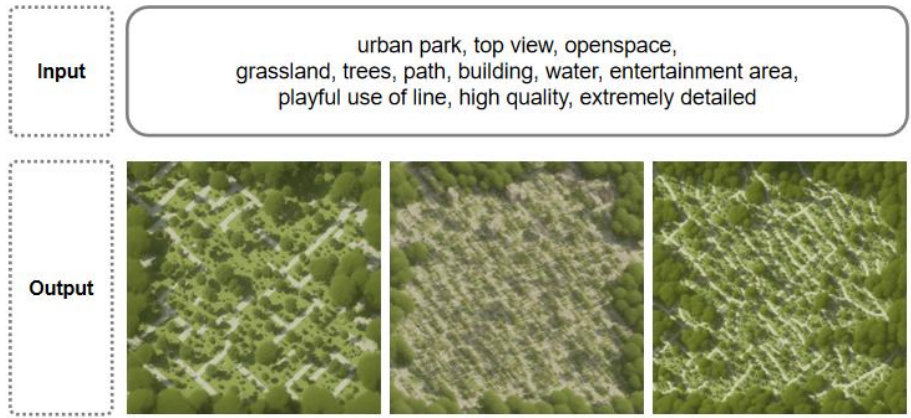

**Fig. 17.** The generation result of only inputting text information in Stable Diffusion

The results of experiment 4(Fig.17) show that the algorithm can hardly generate the park plan, and only generates large areas of trees and complex roads without the essential elements such as "water" and "building" in the description text. It indicates that the algorithm generation results deviate significantly from the general park design scheme, and the existing Stable Diffusion model cannot accurately understand the concept of "landscape" and "urban park".

Therefore, pure text cannot control the existing Stable Diffusion model to generate park design schemes. It is due to the characteristics of landscape design, such as complex design conditions, rich design elements, complex spatial features, and artistic and scientific, which cannot be fully and clearly described by text. Thus, more accurate control is needed to generate design schemes using algorithms.

b. Input layout scheme, semantic confusion of design elements.

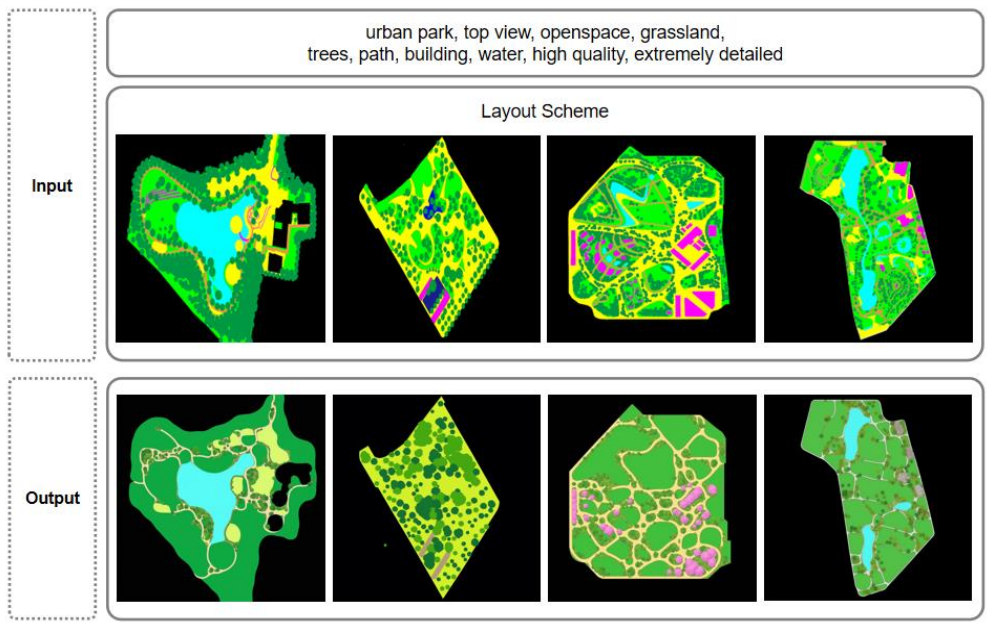

**Fig. 18.** The generation result of inputting text information and park layout in Stable Diffusion.

In the results of experiment 5(Fig.18), Stable Diffusion can place elements according to the shape and position of each color block in the input layout image. However, only water, roads, and green spaces correctly correspond, while buildings and hard paving are mainly generated as green spaces or different colors of trees. Most single trees can be accurately generated, but the algorithm also fails to accurately judge the overlapping tree color blocks as tree groups. Stable Diffusion cannot match the given semantics and design elements individually.

Therefore, controlling Stable Diffusion to generate park design schemes is still impossible by inputting only layout schemes and text descriptions. It may be because Stable Diffusion mainly uses general data for training, such as people, animals, scenery, art, etc., and lacks data sets specifically for the landscape field, resulting in the algorithm's inability to learn professional design rules. There is little training in the field of park design in the current AI model.

c. Input the design scheme generated by small sample training, and the optimization results are apparent.

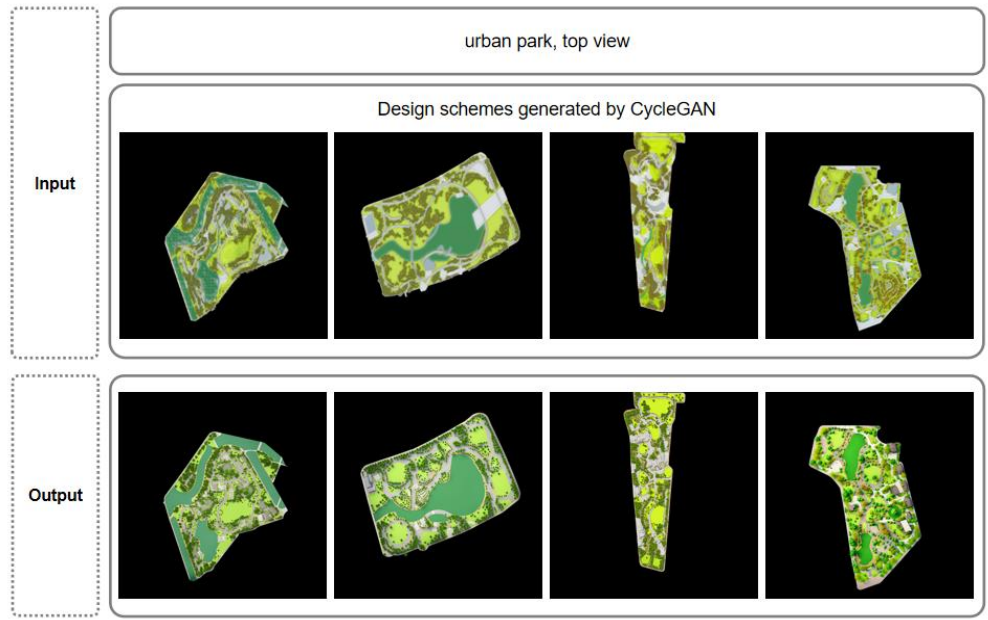

**Fig. 19.** The generation result of inputting text information and park plans generated by CycleGAN in Stable Diffusion

In this experiment, the generated results of Stable Diffusion include rich and diverse site details. They can generate activity sites and vegetation with different contents (Sample 4), and building shadows (Sample 1, 2, 3), and the generated design elements can correspond to the input data individually.

Compared with the layout scheme, the input data of this round of experiments added the color (Sample 1) and texture (Sample 2) of each design element, equivalent to supplementing the corresponding semantic information for each color block. Stable Diffusion algorithm can add more design element details based on clear semantic information. It shows that CycleGAN can obtain the primary design element features and design rules of the park through the landscape-specific data set, which is exactly what Stable Diffusion lacks.

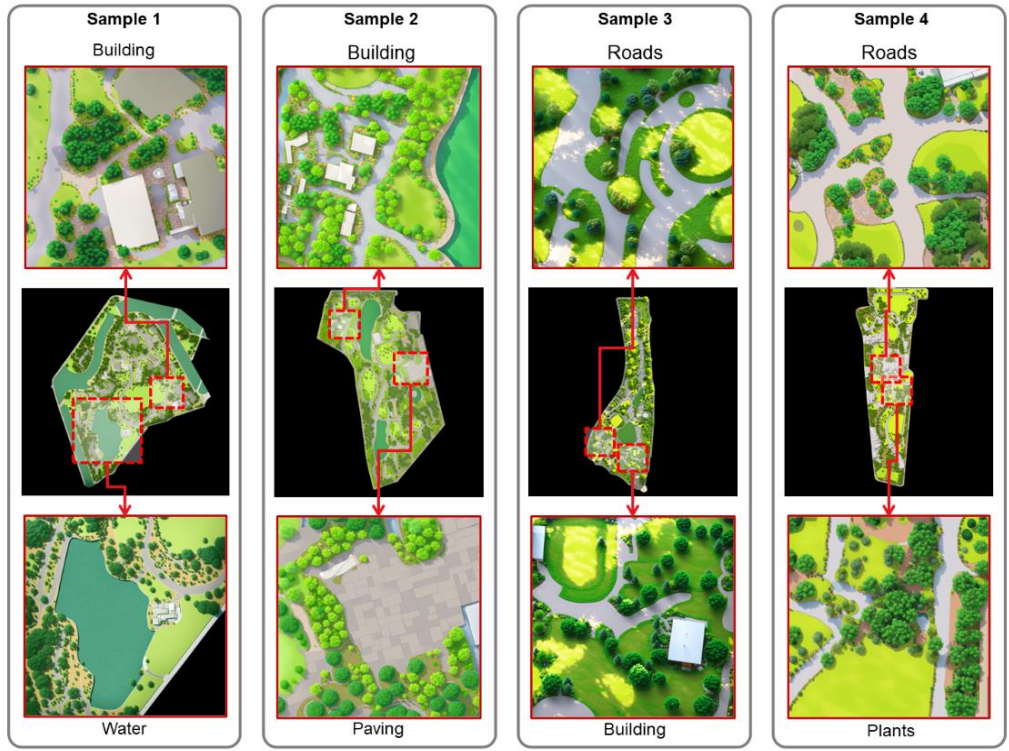

**Fig. 20.** Stable Diffusion optimizes the details of design plans.

# 6.Conclusion

Traditional park design involves designers weighing many constraints and manually converting design information using different software at different stages. With the development of digital technology, algorithm optimization, and data resource enhancement, generative adversarial networks (GANs) are increasingly applied to automatic planning and design in landscape architecture and related fields. However, the scientific problem of generative design is particularly prominent in applying GAN-based landscape planning and design. Based on the latest achievements and defects of landscape architecture generative design, this study proposes a method to enhance the logic and scientificity of generative design application. It establishes a park generative design system of "external environment-layout-scheme". The model generated by the proposed method can mainly assist designers in the initial design stage.

This study applies GANs to three important design links: external environment information extraction, automatic generation of park layout schemes based on external environment conditions, and automatic generation of park design schemes based on layout schemes. It then explores the optimization of generative design drawings and the application of Stable Diffusion in park generative design. We have discussed the main application issues of generative design in three aspects:

(1)Discuss how to constrain generative design. From the perspective of working mode, in the generative design based on deep learning, the designer's participation is low, and the algorithm is too complicated to be effectively constrained. In the current related research, it is possible to quickly generate a complete planning and design scheme based on a few conditions. However, there is little research on generative design considering the external environment. This study introduces external environment information into the park scheme generative design process, making the park scheme generative design closer to the actual design process and improving the scientificity and interpretability of generative design.

(2)Discuss the feasibility of automatic park layout design based on site conditions. This study analyzes the characteristics of the algorithm generation results from the landscape architecture perspective, and evaluates the ability of algorithm generative design. Compared with pix2pix, the park layout generation results of the CycleGAN algorithm are more stable; compared with only inputting the internal conditions of the site, adding the external environment as the constraint condition can make the algorithm more flexible to use various design elements, and consider the impact of the external environment on the park design, making the generation results closer to the actual park layout.

(3)Provide new methods and references for optimizing generative design drawings. This study completes optimizing the park design scheme generated by the algorithm based on Stable Diffusion. It solves the problem that the training effect of some small samples is difficult to improve. At the same time, by comparing the generation results of park plans controlled by different information inputs, the limitations of Stable Diffusion in generative design and the demand for park generative design for targeted professional data sets are obtained.

This study faces many obstacles and limitations, summarized below, and should be addressed in future research. This study initially developed a park generative design method under the constraint of external environment conditions of the site, but only considered land use types as environmental conditions, which are far from the natural landscape architecture design work. For example, it ignored the combination form of the park site status and the complex problems such as site topography, climate, and human flow. Future research should introduce more site design conditions and three-dimensional space conditions by improving the data set and algorithm to conduct generative design research under different constraints and enhance the controllability and interpretability of the design layout scheme. The proposed constraint method can be applied to similar layout generation problems, such as building layout generation and other urban outdoor space layout generation.

Moreover, due to the small sample size of the park data set, some design elements with close colors, complex or diverse textures, such as roads, hard ground and concrete buildings, may have some original color confusion and deviation in the design scheme generated by CycleGAN. Therefore, the data set should be expanded in the future. Furthermore, Stable Diffusion and other AIGC models currently lack data sets targeted at the landscape architecture field as training support. Professional park data sets can also be used for Stable Diffusion model training to improve the design scheme generation effect. The production of related professional data sets in landscape architecture is also crucial to promote generative design in the field of landscape architecture.